\documentclass[10pt,twocolumn,letterpaper]{article}

\usepackage{cvpr}
\usepackage{graphicx, wrapfig, lipsum}
\usepackage{comment}
\usepackage{wrapfig}
\usepackage{amsmath, amssymb} % define this before the line numbering.
\usepackage{breqn}
\usepackage{enumitem}

\newcommand{\EE}{\mathbb{E}}

\usepackage{color}
\usepackage{cite}
\usepackage{booktabs}

\usepackage[utf8]{inputenc}
\usepackage[ruled,vlined]{algorithm2e}
\usepackage{times}
\usepackage{epsfig}
\usepackage{graphicx}
\usepackage{amsmath}
\usepackage{amssymb}
\newcommand{\btheta}{\boldsymbol{\theta}}

\newcommand{\etab}{\boldsymbol{\eta}}
\newcommand{\bomega}{\boldsymbol{\omega}}

\newcommand{\bw}{\mathbf{w}}
\newcommand{\bs}{\mathbf{s}}
\newcommand{\bg}{\mathbf{g}}
\newcommand{\bi}{\mathbf{I}}
\newcommand{\pol}{\pi_{{\theta}}(a_t\,|\,\bs_t,\mathbf{g}_\tau,\btheta)}
\newcommand{\softmax}{\text{softmax}}
\newcommand{\img}{\bs^{\star}}
\newcommand{\balpha}{\boldsymbol{\alpha}}
\newcommand{\sr}{\rho} % {\bar{\sigma}^2}
\usepackage{xcolor}% http://ctan.org/pkg/xcolor/
\usepackage{algpseudocode}% http://ctan.org/pkg/algorithmicx
\algnewcommand{\algorithmicgoto}{\textbf{go to}}%
\algnewcommand{\Goto}[1]{\algorithmicgoto~\ref{#1}}%

\usepackage[pagebackref=true,breaklinks=true,letterpaper=true,colorlinks,bookmarks=false]{hyperref}

\cvprfinalcopy % *** Uncomment this line for the final submission

\def\cvprPaperID{6779} % *** Enter the CVPR Paper ID here

\def\sexyname{ForeSIT\xspace}

\ifcvprfinal\fi
\begin{document}

%%%%%%%%% TITLE
%\title{Imagine Where to Go: Latent Trajectory Imagination for Visual Object-Target Navigation}

\title{Learning for Visual Navigation by Imagining Success}

\author{Mahdi Kazemi Moghaddam, Ehsan Abbasnejad, Qi Wu, Javen Qinfeng shi and Anton Van Den Hengel\\
The Australian Institute for Machine Learning\\ The University of Adelaide\\
{\tt\small mahdi.kazemimoghaddam, ehsan.abbasnejad, qi.wu01, javen.shi and anton.vandenhengel@adelaide.edu.au}
%\and

}

\maketitle
%\thispagestyle{empty}

%%%%%%%%% ABSTRACT
\begin{abstract}
   %Most of the current state-of-the-art visual navigation approaches define the problem as a multi-task reinforcement learning. Learning an optimal policy for this problem that can generalise across tasks to previously unseen environments is a very challenging problem. However, learning an optimal policy that generalises well to previously unseen environments across tasks is very challenging.
       Visual navigation is often cast as a reinforcement learning (RL) problem. Current methods typically result in a sub-optimal policy that learns general obstacle avoidance and search behaviours. For example, in the target-object navigation setting,
       %A (i.e., navigate to find a target object), 
       the policies learnt by traditional methods often fail to complete the task, even when the target is clearly within reach from a human perspective. In order to address this issue we propose to learn to imagine a latent representation of the successful (sub-)goal state. To do so, we have developed a module which we call {Foresight Imagination (\sexyname)}. \sexyname is trained to imagine the recurrent latent representation of a future state that leads to success, e.g. either a sub-goal state that is important to reach before the target, or the goal state itself. By conditioning the policy on the generated imagination during training, our agent learns how to  
       %A ***navigate towards different objects efficiently. 
       use this imagination to achieve its goal robustly.
       %/A*** Efficiency isn't the issue (is it?), it's robustness
       Our agent is able to imagine what the (sub-)goal state may look like (in the latent space) and can learn to navigate towards that state. We develop an efficient learning algorithm to train \sexyname in an on-policy manner and integrate it into our RL objective. The integration is not trivial due to the constantly evolving state representation shared between both the imagination and the policy. We, empirically, observe that our method outperforms the state-of-the-art methods by a large margin in the commonly accepted benchmark AI2THOR environment. Our method can be readily integrated or added to other model-free RL navigation frameworks. 
\end{abstract}

\section{Introduction}
%Einstein: Imagination is more powerful than knowledge.
% Navigation is one the prized problems in computer vision and robotics. Traditional approaches using geometric methods typically 
The ability for an intelligent agent to navigate through an environment to carry out human provided instructions is one of the primary objectives of robotics and artificial intelligence. Visual navigation has emerged to enable such agents to \emph{learn} (\eg using deep learning) to move towards a given target object \cite{cmp}. The problem is then defined as a deep reinforcement learning \cite{thor_target_driven} problem since the agent needs to learn a navigation policy and stopping criteria conditioned on finding a specific target. It is a particularly challenging problem since the agent needs to learn how to avoid obstacles and take actions by distinguishing similar but visually different targets in a complex environment. Furthermore, there are typically multiple action sequences (\ie trajectories) that could lead to a successful navigation at each starting state, let alone the trajectories that might fail. Learning to select the right action at each time step to create a trajectory that leads to the goal is the primary challenge. 

\begin{figure}
    \centering
    \includegraphics[width=75mm]{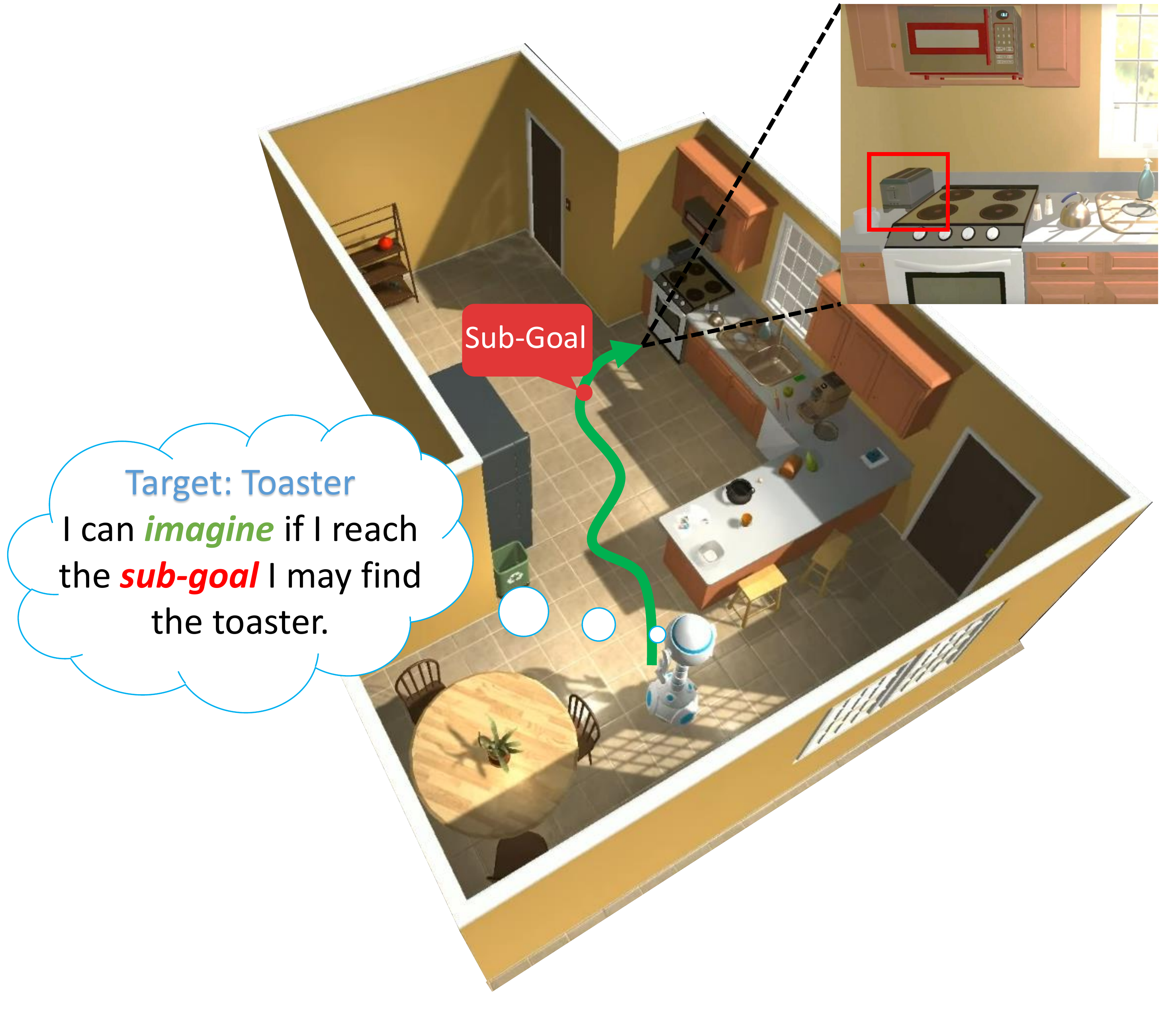}
    \caption{ 
    %A We enable an agent to imagine a sub-goal state through which it's navigation is more successful in an unseen environment. 
    Enabling an agent to imagine states on the path to success improves its ability to carry out complex tasks, particularly in unseen environments. 
    As opposed to the conventional approaches, our \sexyname takes actions not only based on the current state, but also the potential future, to achieve its goal.}
    %A*** THat caption is a bit clumsy.  I'm not sure what it's mean to mean
    \vspace{-10pt}
    \label{fig:fig1}
\end{figure}

%A The challenge can be associated, in particular, to receiving 
Furthermore, in navigation the agent only receives
%/A
a delayed reward, after task completion, hence knows the reasonableness of its actions in \emph{hindsight}. 
%A That means the agent only gets feedback when the decisions are made. 
%A *** The above sentence contradicts the one before
% The agent thus only knows the reasonableness of its actions in \emph{hindsight}. 
%and does not have an indication of its potential success from a given state, other than an internal estimate of its likely accumulative reward (\ie the value function). 
Explicitly incorporating the transition in the environment for better prediction of the potential outcome of the actions, hailed model-based RL, is also developed. However, model-based approaches are generally data hungry and harder to train since every state transition in the environment has to be accurately modelled.
In addition, it is impossible for the agent to identify which specific actions had the greatest impact on its success; for navigation, for instance, it does not know whether avoiding an obstacle, going straight in a hallway, or stopping near a book was the most important decision. This lies at the heart of RL which seeks to solve the problem of credit assignment from delayed environmental feedback.

To mitigate the above mentioned issues we propose our \emph{\underline{F}oresight \underline{S}ub-goal \underline{I}magination uni\underline{T}} (\sexyname). 
%A*** That's a great name!
Intuitively, as shown in Figure~\ref{fig:fig1}, our imagination module generates a representation of a potential successful trajectory given the target and the initial state (\ie the egocentric view of the environment).
%, with a single MLP. 
\sexyname provides the agent with foresight of the future sub-goal through which the agent could successfully achieve its target. In a way our approach alleviates the problem with model-free RL in visual navigation by allowing the agent to incorporate a foresight credit assignment. This represents a significant advantage.
%Therefore, the agent can take actions to minimise the distance to the imagined state in order to maximise its expected reward in the same latent space as the policy. 
% By conditioning \sexyname on the target object (e.g. task) the agent can learn to imagine sub-goal states specific to a given object more efficiently. 

% it is easier to navigate towards a target by having access to a representation of the target itself or the path that leads to that target. This way the agent can continuously compare the new observed states with the state that it is trying to navigate towards and take actions accordingly. 

%furthermore finding the goal state to imagine is difficult which we do using attention

% Training an efficient imagination module to predict a successful sub-goal state in the absence of ground truth state action pairs is highly non-trivial. One issue is that the state representation in RL evolves through training and 

% In order to address this 

Our imagination module helps the agent in multiple ways: first, it provides the agent with an imagined representation of the sub-goal state that will help with successful task completion. Second, it helps the agent exploit its memory of successful traversed states, more efficiently. Thirdly, it helps the agent to perform credit assignment better than conventional heuristic based methods by helping the agent better learn the distribution of the states that lead to higher final reward.

%One of the challenges of target-object navigation is that the goal state observation might sometimes include only a small part of the target object. Therefore, imagining the final image observation only might not provide a strong signal indicating the shape of the target object the agent needs to find. Furthermore, for each given target at a certain location there are many different goal states looking at the object from different positions and angles. Therefore, the reconstruction based methods to learn the observation in the final states do not work efficiently.

%\todo{refine here again}
We train our \sexyname using the successful trajectories collected by the agent and integrate into a navigation policy, in particular an on-policy actor critic algorithm \cite{a3c}. We learn the sub-goals as states that highly correlate with successful navigation using an attention mechanism.
We then train a model to predict the highest attended state representation, signifying its impact on success. That way, even if that single collected trajectory is sub-optimal, our \sexyname module could potentially generate a more optimal sub-goal by learning from other more successful ones. Moreover since our \sexyname is akin to a plugin for RL in a navigating agent, the imagination module and the main policy evolve collaboratively. This is particularly desirable for smooth integration into different RL approaches without extra environments, or algorithm specific requirements. 

We gradually introduce imagination into the policy so that the agent performs sufficient exploration, as well as learning to imagine accurately. Furthermore, we modify the value function to operate on a representation of the trajectory rather than an individual state using an attention-based internal memory.
Practically, we share the state space between our imagination module, and the policy we define, using a deterministic Long Short Term Memory (LSTM). The LSTM encodes a history of the observation and actions in each trajectory. Therefore, imagining sub-goal states in the state representation of the policy is effectively imagining a trajectory leading to success. We find that our simple \sexyname module works well on multiple baseline and state-of-the-art models in the AI2THOR environment \cite{thor_env}.

Our primary contributions are as follows:
%\todo{check again with the contributions; some details and why it is challenging is missing; no mention on the application itself}
\vspace{-5pt}
\begin{enumerate}[noitemsep]
%\vspace{-5pt}
    \item We propose an imagination module \sexyname that is able to generate foresight of states on the path to success, conditioned on the first state and the target object.
    %\vspace{-5pt}
    \item We propose a method to efficiently integrate the \sexyname module into reinforcement learning which does not require pre-exploration for data collection.
    %\vspace{-5pt}
    \item We show the effectiveness of our method in improving visual navigation performance when added to multiple different baselines, including previous state-of-the-art methods.
    
    %\item We introduce foresight imagination for on-policy RL that effectively boosts performance in a near-real-world 3D environment for visual navigation. 
\end{enumerate}

\section{Related Work}
\paragraph{Visual Navigation}
Recently, visual navigation research has gained momentum due to the availability of high quality simulation environments \cite{r2r, thor_env, habitat, gibson}. Different tasks \cite{citi_nav, look_before_leap, hanna, reverie, touchdown, dialog_nav, eqa} and ideas \cite{savn, graph_topo, scene_priors, thor_target_driven} have been explored. Some recent methods use imitation learning \cite{r2r, touchdown, look_before_leap} in simulation environments that provide ground truth state-action pairs. A more promising area of research, however, is to apply RL in the absence of ground truth annotations \cite{savn, thor_target_driven, scene_priors}. Our method introduces imagination for RL-based visual navigation for the first time. Therefore, we compare our method with related works in other tasks and environments.

\vspace{-13pt}
\paragraph{Experience Replay}
Modifying the distribution of trajectories in a replay buffer, known as hindsight experience replay, has been explored in various works \cite{her, imagine_explore, her_goalrl}. In this body of work the core idea is to modify the distribution of goal states in the replay buffer (e.g. hindsight) to train a more generalisable policy able to reach arbitrary goals provided during test time. In contrast, here we introduce a method that is trained in hindsight and tested in foresight to generate successful (sub-)goal states for navigation.  

\begin{figure*}
\centering
\includegraphics[width=0.85\textwidth]{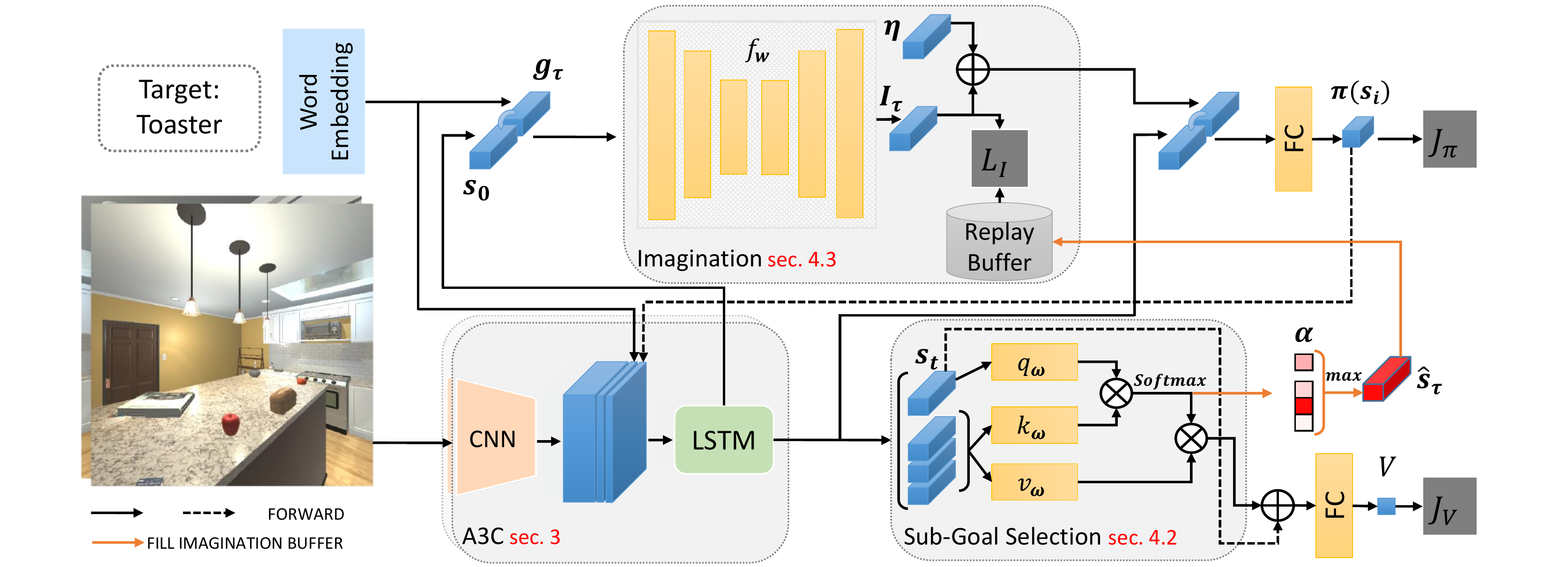}
% \vspace{-3pt}
\caption{Our \sexyname module efficiently augments actor-critic RL and is trained along with the policy sharing the state representation. We use a dot-product attention mechanism to identify the sub-goal state that leads to maximum reward and train our \sexyname module to generate that state.}\label{overview_fig}
\vspace{-10pt}
\end{figure*} 

\vspace{-13pt}
\paragraph{Goal-Conditioned RL}
Our method is also closely related to the area of goal-conditioned RL where the policy is conditioned on a given goal state. In \cite{leap} the authors use combinatorial optimisation on the states encoded using a Variational Auto-Encoder \cite{vae} to find a set of reachable sub-goals along a past trajectory. They use the selected sub-goals to optimise the policy using Q-learning \cite{dqn}, an off-policy RL algorithm. In \cite{rig} the authors use self-imagined goals to train the agent to reach to arbitrary goals. Most recently, in \cite{ifig} the authors use a VAE to learn to generate goals during test time. While these ideas are related to that proposed here, there are important differences: in contrast to \cite{ifig} we train our method along with an on-policy RL algorithm; the constantly evolving shared state representation renders the task very challenging. In \cite{ifig} the training and testing environment are simple and similar in distribution, limited to moving pucks. Finally, in their method only the final state is generated as the target while we use an attention mechanism to identify the optimal sub-goal state.

\vspace{-13pt}
\paragraph{Model-Based RL}
Our \sexyname module generates future state representations, and thus is related to model-based RL. In \cite{i2a} the authors train a recurrent model of the environment dynamics that can be used for model-based planning by unrolling a trajectory in the imagination. In a similar but more recent method \cite{dreamer} the authors train the policy purely on imagined trajectories. In all of these methods the environment model is a recurrent reconstruction-based model inspired by \cite{world_models}. These state encoders are not directly trained for policy optimisation and thus can be sub-optimal. Moreover, the mentioned works have only been tested on toy environments and the generalisation to real 3D tasks is a big question. In contrast, we enable imagination for more challenging on-policy RL in complex navigation environments.

\section{Background} \label{setup}
In visual navigation the objective for an agent starting from a random initial location is to take a sequence of movement actions (collectively, a trajectory) to reach a named target object using only observed ego-centric RGB visual input from the environment. An episode of this navigation trajectory ends if either the agent takes the stop action or the maximum number of permitted actions is achieved. A successful trajectory sees the agent stop within a defined circular proximity of any instance of the target object. This problem is conventionally defined as a Partially-Observable Markov Decision Process (POMDP) denoted by the tuple $\{O, S, A, \mathcal{G}, \mathcal{P}, r, \gamma\}$ \cite{learn_to_nav,savn}. Here, $O$ is the space of visual observations, $S$ is the state space as encoded and observed internally by the agent, $A$ is the action space, $\mathcal{G}$ is the set of target objects given by the environment, $\mathcal{P}:=p(\bs_t | \bs_{t-1},a_{t-1})$ is the transition function or environment dynamics model (unknown) for state $\bs_t\in S$, and $r$ is the reward function and $\gamma$ is the reward discount factor. 

Formally, a trajectory $\tau$ of length $T+1$ consists of a tuple $(\bs_0, a_0, r_0; \bs_1, a_1, r_1; ..., \bs_T, a_T, r_T)$ that is generated by taking action $a_t$ at time $t$ and observing the next state according to the dynamics of the environment $\bs_{t+1} \sim p (\bs_{t+1} \,|\, \bs_{t},a_{t})$. A reward $r_t = r(\bs_t, a_t, \bs_{t+1})$ is received from the environment at each time step, which is conventionally \cite{thor_target_driven, savn} defined as a large positive number for a successful trajectory  and a negative penalty otherwise. 

A common approach is to use actor-critic RL methods for learning the optimal policy\footnote{Here, we drop the explicit dependency on the visual observations since our state $\bs_t$ is learnt using a (recurrent) deep neural network that captures its representation.} \cite{thor_target_driven, scene_priors, savn} for navigation $\pol$ to choose action $a_t$ at time $t$ conditioned on an embedding of the target object's name $\mathbf{g}_\tau$ (e.g. using GloVe embedding \cite{glove}) for a given environment (\ie a room in AI2Thor) $E$. We use $\btheta$ to denote the set of all the parameters for RL. Learning involves minimising $\mathcal{J}_{\pi}(a_t \,|\, \bs_t,\btheta)$ the negative of expected advantage function while
% defined as $A_\theta(\bs_t, a_t) = r_t + \gamma V_\theta(\bs_{t+1}) - V_\theta(\bs_t)$ 
minimising the difference of the estimated value function and the true return
$\mathcal{J}_{V}(\bs_t,\btheta)$ where we have:
% . As such, in the actor-critic models using the following two alternating objectives:\todo{one is max and another is min, right?}
\begin{align}
\mathcal{J}_{\pi}(a_t \,|\, \bs_t,\btheta)  &= -\log\pi (a_t\,|\,\bs_t,  \bg_\tau; \btheta)(r_t\,+\label{a3c_objective_1}\\ 
&\qquad\,\gamma V_\theta(\bs_{t+1}) - V_\theta(\bs_t)) + \beta_H  H_t(\pi)  \nonumber\\
\mathcal{J}_{V}(\bs_t,\btheta) &= \frac{1}{2}( V_\theta(\bs_t) - R)^2 
\label{a3c_objective_2}
\end{align}
and $R = \EE_{\tau \sim \pi} [\sum_{i = t}^{T} \gamma^t r_i\, |\, \bs_t]$. Here, $H_t$ is the entropy of the policy that acts as a regularizer with $\beta_H$ as its hyper-parameter.
% \begin{align}
% \mathcal{J}_{\pi}(a_t | \bs_t,\btheta)  &\approx -\log\pi (a_t|\bs_t,  \bg_\tau, \bi_\tau; \btheta)(r_t(a_t | \bs_t) \nonumber\\ 
% & \qquad\,+\gamma V(\bs_{t+1}) - V(\bs_t)) + \beta_H  H_t(\pi) \label{a3c_objective_1} \\
% \mathcal{J}_{V}(\bs_t) &= \frac{1}{2}( V(\bs_t) - R)^2 
% \label{a3c_objective_2}
% \end{align}
The value function acts as a critic for the policy's generated actions. Typically random samples from the environments containing various room types are taken by rolling out the policy to obtain state-action-reward tuples to compute these terms.
Note that the value function is the only information the agent has about the future potential of the current state during training. However, during the test time, the value function is not used by the agent any more.
% In the subsequent section we discuss how to empower our agent with a foresight imagination for estimating the potential future states leading to improved success in visual navigation. 

%\todo{It has been shown in \cite{wacv} that more accurate value estimation in the critic helps with training a more optimal policy. This is because the critic provides signals for better credit assignment to the taken actions based on the advantage function defined as $A(s_t, a_t) = r_t + \gamma V(s_{t+1}) - V(s_t)$ where $V(s_t) = \EE_{\tau \sim \pi} [\sum_{i = t}^{T} \gamma^t r_i | s_t]$. This is important as in the next section we use the learned correlation over the past states collected in an internal memory to find prominent states observed during a sample trajectory. We propose to use multi-head attention mechanism \cite{} in a novel way to perform more accurate value estimation.}

% \section{Method}
\section{Foresight Sub-goal Imagination}
% In this section we first present the problem definition and then introduce our \sexyname module and explain how it is integrated into our backbone RL algorithm.

In this section we detail our approach to providing RL with explicit foresight, or imagination, for navigation. 
%\vspace{-15pt}

%\subsection{Imagination Module}
\subsection{Imagination Module Overview}
Having a single policy with which to achieve various visual navigation tasks based solely on a target object's name is non-trivial. % We introduce our \sexyname module into the POMDP that helps better handling the different target objects by providing the agent with an explicit foresight imagination of the future sub-goal state.
%we redefine the problem as the tuple of $\{O, S, A, I, G, \mathcal{P}, r, \gamma\}$. Note that we have two extra sets of variables, $I$ the set of imagined sub-goal states using our proposed imagination module and $G$ the set of tasks (\eg target objects) given by the environment.
% Our intuition is that as in humans that, a sub-goal or a prominent state is imagined when before navigating for the target, we quantify $\img$
% $$p(\bs_t\,|\,g)= \sum_i p(\bs_t\,|\,\bs_i,g)$$
% $$\bs_g\sim p(\bs_{t+1},\,\bs_{t+2}\ldots\,|\,\bs_t, g)$$
% we can write
% $$p(\bs_{t+1},\,\bs_{t+2}\ldots\,|\,\bs_t, g)\propto p(\bs_t\,|\,\bs_{t+1},\,\bs_{t+2}\ldots g)\cdotp p(\bs_{t+1},\,\bs_{t+2},\ldots)$$
% $$p(\bs_{t+1},\,\bs_{t+2}\ldots\,|\,\bs_t, g)\propto p(\bs_t\,|\,\bs_{t+1},\,\bs_{t+2}\ldots g)\cdotp p(\bs_{t+1},\,\bs_{t+2},\ldots)$$
The aim of our \emph{Foresight Sub-goal Imagination} (\sexyname) module (see Figure \ref{overview_fig}) is to provide the agent with an imagined sub-goal state\footnote{Note that a sub-goal state could potentially be the final goal state.} such that it is a future state through which the navigation has maximal chance of success. Incorporating such a sub-goal in the policy enables the agent to take better actions since it has an indication of what likely states to visit and eliminates a need for unnecessary exploration at test time.
There are two main challenges to overcome: (1) determining the sub-goal state to be learned% by reviewing the observed trajectory in hindsight
, and (2) imagining or generating that sub-goal given the initial state and target object in an unseen environment. 

In determining the sub-goal state to be learned, we  note that the sub-goal most valuably imagined is that with the highest impact towards receiving the maximum reward. In other words, the optimal sub-goal is the one with which we could learn the best value function to estimate the expected reward. Intuitively, when the agent reaches the optimal sub-goal, finding the target and receiving the maximum reward should be easy (see Section \ref{attention}).

Once we find the optimal sub-goal, we can train a model to predict or \emph{imagine} an instance for an unseen environment.  
% one using a generative model conditioned on the initial state and the target. 
To that end, we collect a replay buffer of the successful trajectories to train our \sexyname. Intuitively, when an agent navigates through particular states to achieve its target, imagination of the sub-goal from a related successful trajectory helps the agent to identify how to plan and take actions  in unseen environments (see Section \ref{train_img}). This imagination is conditioned on the initial state and the target, and is integrated into a parallel execution of the policy in Asynchronous Advantage Actor-Critic (A3C) \cite{a3c} for best performance. This allows the agent to learn to both imagine and navigate (see Section \ref{sec:IMac}).
\subsection{Sub-goal Selection} \label{attention}
In the first step we consider learning to identify the sub-goal state through which a navigation trajectory is successful. To that end, the  agent has to consider its current state and the sub-goals it has navigated through with their corresponding visual, target and dynamics representation.  Intuitively, for the agent to obtain a high reward in a straight path through a hallway, for instance, it should have chosen the best action while at the previous corner to be successful. Since the only aspect of the RL that considers the future reward is the value function, we modify Eq.~(\ref{a3c_objective_1}) using a residual function of the past states as:
\begin{align}
\small
&V_\theta(\bs_t)  \approx V_\theta(\img_t),\qquad \img_t = \sum_{j=0}^{t}\alpha_jv_\bomega(\bs_j)+\bs_t\,.\label{eq:att_val}
\end{align}
%where $s_t$ is the state at timestep $t$. 
Here, $v_\bomega(\bs_j)$ is a linear function of the input and $\alpha_j$ is the $j$th dimension of $\balpha$, which is defined as follows:
\begin{align}
\small
\balpha = \softmax\left(\frac{q_\bomega(\bs_t)k_\bomega([\bs_{0}: \bs_{t}])^\top}{\sqrt{t+1}}\right).
\label{eq:att}
\end{align}
%\todo{are you sure it is correct? you removed any dependence on the past states now and no need to introduce s_o:t}
Here, $q_\bomega$ and $k_\bomega$ are linear functions analogous to the query and key in an attention mechanism \cite{transformer} with $\bs_{0:t}$ the concatenation of the states up to time $t$. We denote all of our sub-goal selection parameters by the set $\bomega$. Moreover, $\alpha_j$ is the correlation between state $j$ and the current state $t$, and its magnitude specifies the likelihood that state $j$ is an important sub-goal to reach. 
Effectively, we use the attention mechanism described above to identify the sub-goal state that minimises the state value function estimation error.
%A*** , and hence the estimation of the future success. 
%A*** I had to delete the above because it said you want to minimise future success, which seems unlikely %E agreed
Using the key-query product above we choose the state in the past that is most related with the current state. Interestingly, if the state is novel (\ie uncorrelated with the past states) we allow the agent to assign a high attention weight to its most recent state.
%A*** "current with higher weight." Doesn't make sense  %E hopefully fixed
We select the state with the maximum attention weight as the sub-goal state at the end of a trajectory, i.e. $\hat{\bs}_\tau = \bs_{t^*}$ where $t^*=\arg\max_t\alpha_t$ (note $\balpha$ has length $T+1$). We subsequently task \sexyname to learn to generate this sub-goal. Using this method we ensure that the imagination will guide the policy towards a state that has the highest correlation to a successful goal state. It is also easy to integrate into the existing RL approach. 
% Our proposed method for sub-goal selection has two major benefits: first it encourages taking more optimal trajectories in case a successful trajectory is longer than optimal. Second, it ensures the imagined sub-goal is reachable since it is selected from a real trajectory rather.

% Finally the optimal sub-goal is selected as the $\hat{\bs}_\tau = \bs_j$ where $j$ is the index of the maximum value in the attention vector $\balpha$.
%\subsection{Imagine and Learn to Imagine} \label{train_img}
\subsection{Learning to Imagine} \label{train_img}

For learning to imagine, or generate, the selected sub-goal state $\hat{\bs}_\tau$ we consider a replay buffer. The replay buffer, denoted by $M$ is filled with tuples of $(\bs_0,\bg_{\tau}, \hat{\bs}_\tau)$, an initial state, an embedding representation of the target object $\mathbf{g}_\tau$ for trajectory $\tau$ and the sub-goal. We then devise the following objective to train an imagination function $f_\bw$:
% \begin{equation}
%         I_\tau = \tanh(\boldsymbol{W}_{I} [\bs_0: g_\tau])
% \end{equation}
\begin{equation}
\small
    \min_\bw\quad \mathbb{E}_{(\bs_0,\bg_{\tau}, \hat{\bs}_\tau)\sim M}\,\,\,\Big| \hat{\bs}_t\, -\, f_\bw([\bs_0: \mathbf{g}_\tau])\Big|
\label{imagine_loss}
\end{equation} % I_\tau = \tanh(\boldsymbol{W}_{I} [s_0: g_\tau])
where $\bw$ is the set of parameters of our \sexyname and $[\,:\,]$ denotes concatenation of vectors. For $f_\bw$ we use a multi-layer percepteron and a bottleneck with the intuition that the structure of the sub-goal distributions lies in a lower dimensional space. Intuitively, the imagination module does not need to generate every single dimension of the state accurately, since there might be unnecessary information about the other objects or the background scene in the representation. It might also include dynamic objects that constantly change location. Using a smooth version of an L1 loss, we avoid penalising such inexact predictions too harshly. 

We use a shared recurrent state encoding for both the policy and the \sexyname module. Therefore, the state representation generated by the \sexyname module not only has the information about the sub-goal state's visual observation, but also encodes the history of the past observation and action pairs before that state. Hence, \sexyname essentially generates a representation of the whole trajectory that leads the agent to a successful goal state. Furthermore, sharing the state encoder that is trained along with the policy helps generate goal-states that are directly useful for the action selection. This is as compared to methods such as \cite{world_models}  where the agent needs to first perform random policy search to collect images from the environment, and then use only the visual features for generating future states. 
%A*** "Visual features alone are hardly sufficient." This doesn't add anything but opinion. If it's important then it needs to be more convincing.

% This is a major superiority of our method compared to previous imagination works such as \cite{world_models} where the state encoding is done using a reconstruction-based model. Those methods generally require a pre-training phase and the learnt representations are not directly optimised for action selection, hence can be sub-optimal.

Since computing the expectation in Eq. (\ref{imagine_loss}) is impractical due to the buffer size and the constant change in the distribution of states, we only consider the latest collected tuples to update the model's parameters. That is, we update the model when $|M|= M_\max$ and empty the buffer.

\subsection{Imagination for Actor-Critic models}
\label{sec:IMac}

To use \sexyname we condition the policy on the imagined sub-goal state. This way the policy learns to take actions based on the current state and the imagined one to achieve its goal. %The metric is learnt during the policy optimisation as the \sexyname module evolves.
One potential issue with this approach is that the policy could bias towards only exploiting the known imagined sub-goal states and avoid exploration. This could lead to the agent's policy collapsing to a deterministic one which is highly undesirable. We address this issue by adding Gaussian noise to the imagined states. Hence we have the revised policy:
% Therefore, at each time step we sample an action from $a_t \sim \pi_\theta(\bs_t,  \bg_\tau, \bi_\tau)$ where $\bi_\tau=f_\bw([\bs_0: \bg_\tau])$.
\begin{align}
         a_t & \sim \pi_\theta(a_t\,|\,\bs_t,  \bg_\tau, \bi_\tau) \quad\text{and,}\nonumber \\
         \bi_\tau& =f_\bw([\bs_0: \bg_\tau])+\etab, \quad\etab \sim \mathcal{N}(\boldsymbol0, \sigma^2\boldsymbol{I})
         \label{eq:i_tau}
\end{align}
Here, $\boldsymbol0$ is a vector of all zeros, $\boldsymbol{I}$ is the identity matrix, and $\sigma^2$ is the variance of the noise. The additive noise is decayed during training to allow more exploration when the agent is more unsuccessful (\ie the imagination is not robust enough) and exploit otherwise. We choose the noise level by adjusting the variance, that is,
\begin{align}
    \sigma^2 = \max({\sigma}^2_\max - \sr, 0)\
\label{noise}
\end{align}
where $\sigma^2_\max$ is a pre-defined maximum variance threshold, and $\sr$ is a moving average of the success rate over the past episodes. This simple heuristic ensures the noise level is proportionate to the success rate, for instance, if $\sr=0.9$ and success rate is around 90\% we completely remove the added noise.

Finally, we integrate the sub-goal selection and the imagination into  the on-policy Asynchronous Advantage Actor-Critic (A3C) \cite{a3c} algorithm which allows for efficient and parallel training of multiple agents. We devise the following revised objectives which integrates our \sexyname into RL: 
\begin{align}
\mathcal{J}^\star_{\pi}(a_t \,|\, \bs_t,\btheta)  &= -\log\pi (a_t\,|\,\bs_t,  \bg_\tau, \bi_\tau; \btheta)(r_t\,+\label{a3c_objective_1_new}\\ 
&\qquad\,\gamma V_\theta(\img_{t+1}) - V_\theta(\img_t)) + \beta_H  H_t(\pi)  \nonumber\\
\mathcal{J}^\star_{V}(\bs_t,\btheta) &= \frac{1}{2}( V_\theta(\img_t) - R)^2 
\label{a3c_objective_2_new}
\end{align}
We summarise the training of our approach in Algorithm~\ref{train_alg}. As may be observed, in the spirit of A3C, each agent keeps its own buffer and computes its success rate. 

% In our actor-critic RL framework we do not assume access to the ground truth goal states nor we assume a pre-training phase availability. Therefore, we train our \sexyname module along with the navigation policy using our on-policy replay buffer. We start training the policy using maximum entropy exploration and collect the successful trajectory tuples of $(\bs_0, \hat{\bs}_\tau, \mathbf{g}_{\tau})$, as discussed in \ref{train_img}, in a replay buffer, separate for each agent. Then, iteratively, we train each agent's imagination and transfer the learned parameters to the shared model among the agents. This way we enable sharing optimising the state representation directly using the policy which is more efficient. 

%One of the main challenges of our methods is to efficiently train the imagination module and incorporate it into on-policy actor-critic RL algorithm. We consider the more realistic scenario where we train the state encoder and the imagination along with the policy without the need for pre-training or separate pre-exploration.

\begin{algorithm}[t]
\SetAlgoLined
Randomly initialise $\boldsymbol{\theta}, \mathbf{w}, \boldsymbol{\omega}$ \\
Initialise replay buffer $M$ = $\varnothing$ \\
$\sigma^2 = \sigma^2_\max$\Comment{\small{Imagination Noise}} \\
 \While{$\text{episode} <\text{MAX\_EPISODE}$}{
 $(\bs_0, g_\tau)\sim E_{\text{RND}}$\Comment{\small{$E_{\text{RND}}$ is a random environment}} \\
 %$\bs_0 \sim S$\\
 %$\bg_\tau \sim G$\\
 $\bi_\tau = f_\bw([\bs_0: \bg_\tau]+ \etab,\,\, \etab \sim \mathcal{N}(\boldsymbol0, \sigma^2\boldsymbol{I})$ \Comment{Eq.~(\ref{eq:i_tau})}\\
 \While{ $a_t \neq \text{STOP} \And t \leq T$}{
    $a_t \sim \pi_\theta(a\,|\,\bs_t, \bg_\tau, \bi_\tau)$
  }
  %\todo{$T_{\max}$ is the length of episode, right?}\\
  Compute $\balpha$ using eq. (\ref{eq:att}) for the trajectory \\
  Update $\btheta$ and $\bomega$ via eq. (\ref{a3c_objective_1_new}) and (\ref{a3c_objective_2_new})\\
  \uIf{\text{trajectory is successful in the environment}}{ 
%   Get $\hat{\bs}_\tau$ according to \ref{attention}\\
  $\hat{\bs}_\tau=\bs_{t^*}, \,\, t^*=\arg\max_t\alpha_t$ \\
  $M = M \cup \{\bs_{0}, \bg_\tau, \hat{\bs}_\tau \}$ \Comment{\small{Update Buffer}} \\
  }
  Update the average success rate $\sr$ \\
  \uIf{$|M|= M_{\max}$}{\For{$\text{epoch} \leq \text{Epoch}_{\max}$}{
    \For{$\{\bs_{0}, \bg_\tau, \hat{\bs}_\tau \} \in M$}{
   $\bomega \xleftarrow{} \bomega - \beta \nabla_\bw \big| \hat{\bs}_t\, -\, f_\bw([\bs_0: \mathbf{g}_\tau])\big|$}
   }
   $M$ = $\varnothing$
   }
  Update~ $\sigma^2$\Comment{Eq. (\ref{noise})} \\ % \tcp*[f]
  } 
  \caption{Training One \sexyname Agent}
\label{train_alg}
\end{algorithm}

% Balancing the exploration and exploitation is a major challenge in RL, in general. In our method the agent can be tempted to exploit more of the known imagined sub-goal states than exploring new goals. We address this issue by adding a Gaussian noise to the imagined states. The additive noise is decayed through the training to allow more exploitation of the known sub-goals and less exploration as the training progresses:
% \begin{align}
%         a_t \sim \pi_\theta(a_t\,|\,\bs_t,  \bg_\tau, I_\tau+\etab), \quad
%         \etab \sim \mathcal{N}(\boldsymbol0, \sigma^2\boldsymbol{I}).
% \end{align}
% Here, $\boldsymbol0$ is the vector of all zeros, $\boldsymbol{I}$ is the identity matrix and $\sigma^2$ is the variance of the noise.
% The above noisy imagination method, additionally, balances between the determinism and stochastisity of the imagined states. Since our \sexyname module generates the sub-goal states using a deterministic model adding the noise allows for stochastic imagination which encourages exploration. We gradually reduce the variance to zero using the following formulation:
% \begin{align}
%     \sigma^2 = \max({\sigma}^2_\max - \sr, 0)
% \label{noise}
% \end{align}
% where $\sigma^2_\max$ is a pre-defined maximum variance threshold, $\sr$ is a moving average of the success rate over the past episodes. We summarise the training of our approach in Algorithm \ref{train_alg}. As is observed, in the spirit of A3C each agent keeps its own buffer. 

\begin{table*}[t]
\centering
\scalebox{0.9}{
\begin{tabular}{l|l|l|l|l}
\toprule
\textbf{Method}       & \textbf{SPL} & \textbf{SR} & \textbf{SPL\textgreater 5} & \textbf{SR\textgreater 5} \\ 
\toprule
\multicolumn{5}{c}{\textbf{First Setup}} \\
\bottomrule
\textbf{A3C \cite{savn}}          & 14.68        & 33.04            & 11.69                       & 21.44                           \\ 
\textbf{A3C+MAML} \cite{savn}        & 16.15 \small{$\pm 0.5$}        & 40.86 \small{$\pm 1.2$}            & 13.91 \small{$\pm 0.5$}                       & 28.70 \small{$\pm 1.5$}                           \\ \hline
\textbf{A3C+\sexyname}         &\textbf{15.23} \small{$\pm 0.4$}         & \textbf{36.80} \small{$\pm 1.1$}            & \textbf{13.14} \small{$\pm 0.3$}                       & \textbf{27.55} \small{$\pm 1.4$}                           \\ 
\textbf{A3C+MAML+\sexyname}         & \textbf{16.75} \small{$\pm 0.5$}       & \textbf{45.5} \small{$\pm 1.0$}           & \textbf{15.8} \small{$\pm 0.6$}                      &\textbf{34.7} \small{$\pm 1.1$}                           \\ 
\toprule
\multicolumn{5}{c}{\textbf{Second Setup}} \\
\bottomrule

\textbf{A3C+ORG} \cite{tpn_anu}        & 37.5         & 65.3             & 36.1                        & 54.8                            \\ \hline
\textbf{A3C+ORG+\sexyname}         &\textbf{39.41} \small{$\pm 0.3$}         & \textbf{68.0} \small{$\pm 0.6$}            & \textbf{36.85} \small{$\pm 0.4$}                       & \textbf{56.11} \small{$\pm 0.8$}                           \\ 

\bottomrule

\end{tabular}
\vspace{-2pt}
}
\caption{A quantitative comparison of our method against four baseline methods on two different environment setups. \textit{SPL\textgreater 5} and \textit{SR\textgreater 5} show the metrics for trajectories longer than 5 steps. Our approach improves all the commonly used evaluation metrics compared to two previous state-of-the-art methods \cite{savn, tpn_anu}. The two methods are significantly different in the methods used for modelling the state space.}\label{tab:results_compare}
\vspace{-4pt}
\end{table*}

\section{Experiments}
In this section we present implementation details of our method as well as extensive experiments to show how our \sexyname improves the visual navigation performance of multiple significantly different baselines.

\subsection{Experimental Setup}\label{env_setup}

We use the AI2THOR~\cite{thor_env} environment to benchmark our experiments. This simulator consists of photo-realistic indoor environments (\eg houses) categorised into four different room types: kitchen, bedroom, bathroom and living room. We run our experiments on two distinct setups of this simulator for fair comparison against previous state-of-the-art approaches. In both of these setups 20 different scene layouts of each room type are used for training; 5 scenes for validation and 5 for test. We provide the final results on the test set based on the best performing model on the validation set. For fair comparison, we follow the exact object configuration and target object list according to the baseline methods.

In the first setup we follow the configuration as used in~\cite{savn, scene_priors}. In this setup the target object is selected from the following list: pillow, laptop, television, garbage can, box, bowl, toaster, microwave, fridge, coffee maker, garbage can, plant, lamp, book, alarm clock, sink, toilet paper, soap bottle and light switch. Here a trajectory is considered successful if the agent stops within \textbf{1 meter} circular proximity of the target object while the object is visible in the ego-centric view of the agent.

In the second setup we follow the configuration recently introduced in \cite{tpn_anu}. The distribution of the target objects and the object location configuration in this setup is significantly different. The following objects are added to the list of the targets compared to the first setup: cellphone, chair, desk lamp, floor lamp, kettle, pan, plate, pot, remote control, stove burner; moreover, the following objects are removed form the list in the first setup: pillow, box, plant, lamp, toilet paper, soap bottle\footnote{We think due to the incorporation of an off-the-shelf object detector the list of objects in this setup is adjusted by the authors in \cite{tpn_anu} for more efficient detection.}. Lastly, here the successful trajectory criterion is relaxed to a \textbf{1.5 meters} circular distance around the target object.

Following the recent conventions in visual navigation tasks \cite{vln, savn, scene_priors, tpn_anu} we evaluate the performance of our method using two main metrics: Success Rate (SR) and Success weighted by inverse Path Length (SPL), calculated as $\sum s_i \frac{L^g_i}{L_i}$, where $L^g_i$ is the ground truth shortest path to the target, $L_i$ is the length of the trajectory as taken by the agent and $s_i$ is the binary success indicator.

\subsection{Implementation Details} \label{implementation}
%\todo{this is done using a replay buffer, how large is the buffer and how do you update it? do you need to through out the old ones?}
%\todo{add noise details}
We use A3C \cite{a3c} as the basis of our method. This actor-critic algorithm is a good fit for visual navigation task since the agents can explore in parallel and asynchronously, which is more computationally efficient. Our method can be readily integrated into any actor-critic method.

Our backbone model comprises of a single layer LSTM state encoder with 512 hidden states. The input to the encoder at each time step is the visual features extracted from a pre-trained ResNet-18 \cite{resnet} and the Glove embedding \cite{glove} of the target object, as visualised in Figure \ref{overview_fig}. Note the Glove embedding in the second setup is replaced with a one-hot vector embedding following the settings in \cite{tpn_anu}. The policy comprises of a single fully-connected layer that outputs the distribution over 6 actions, \textit{\{RotateLeft, RotateRight, MoveAhead, LookDown, LookUp, Stop\}}, using a Softmax activation function. The state-value head is a two layer MLP that maps the attended state representation to a single scalar value. We use a reward of 5 for task completion and a negative step penalty of -0.01 for each taken action. We implement the attention mechanism as three fully-connected layers of size 512 that map the query, key and value. 

Our \sexyname's $f_\bw$ comprises 6 fully-connected layers that receive the concatenation of the state representation and the target object embedding as  input. We use a $\tanh$ non-linearity in all layers to mimic the behaviour of the LSTM state encoder. 

We use a replay buffer of size 32 for each agent to train its imagination separately. After the imagination module is trained, the weights are transferred to a model shared  between the agents and the replay buffer emptied.
For imagination noise we use $\sigma^2_{max} = 0.9$ and the last 100 episodes to compute the moving average for the success rate.

\subsection{State-of-the-art Results}

In Table \ref{tab:results_compare} we compare our approach to that of state-of-the-art baselines in two separate simulator setups as discussed in Section \ref{env_setup}. The first baseline, \textbf{A3C}, is only using the backbone model described in Section~\ref{implementation} trained using A3C RL algorithm. This is the main baseline that shows how a simple RL objective can perform without any extra components or modifications. \textbf{A3C+\sexyname} is the variant of A3C with our \sexyname module added. We see that empowering the agent with imagination improves the success rate by more than 3\%. The improvement on longer-horizon tasks, \eg longer trajectories, is even more significant, more than 6\%. We conjecture that this is mainly because in longer trajectories the agent can forget the objective by focusing the resources on encoding the complex visual observations, or it is more likely to follow a trajectory that does not lead to the target. Adding our \sexyname module, however, helps the agent both remember the target as well as taking a more confident trajectory by aiming for a sub-goal.%\todo{is it a claim based on the experiments or just a motivation?} 

\begin{table}[t]
\centering
\scalebox{0.71}{
\begin{tabular}{l|c|c|c|c}
\toprule
\textbf{Method}      & \textbf{Bathroom} & \textbf{Bedroom} & \textbf{Kitchen} & \textbf{Living} \\ 
\toprule
\multicolumn{5}{c}{\textbf{First Setup}} \\
\bottomrule
\textbf{ +MAML}  & \textbf{28.49}/69.6      & 8.65/\textbf{29.2}    & 17.8/43.6     & 7.71/21.6            \\ \hline
\textbf{ +MAML+\sexyname}  & 27.03/\textbf{73.6}     &\textbf{8.81}/27.6      & \textbf{21.55/54.0}     & \textbf{9.61/26.8}  \\ \hline
\toprule
\multicolumn{5}{c}{\textbf{Second Setup}} \\
\bottomrule
\textbf{ +ORG}  & \textbf{49.87/83.89}      & 35.43/62.21    & 38.63/69.02     & 29.33/47.83            \\ \hline
\textbf{ +ORG+\sexyname}  & 47.44/80.4      & \textbf{38.15/65.6}    & \textbf{41.09/72.87}     & \textbf{30.39/52.14}            \\ \hline
\toprule
\end{tabular}
}
%\vspace{-5mm}
\label{tab:per_room}
\caption{Detailed comparison with against previous state-of-the-art methods on the two different setups; SPL/SR are reported per room type. Our method is general enough to improve performance in 3/4 of the room types, with marginal performance impact on 1/4. Notably, we improve the SR of trajectories in the "kitchen" and "living room" by a large margin where the trajectories are generally longer and foresight is more important.}
\vspace{-3mm}
\end{table}

To evaluate the performance of our approach when updating the model in the test environment we utilise meta-learning similar to \textbf{A3C+MAML} \cite{savn}. This allows the agent to adapt to  unseen environments at test time.
%The second powerful baseline that we compare our model with is named \textbf{A3C+MAML} \cite{savn}. This method is the previous state-of-the-art method in the widely accepted first simulator setup. In that method \cite{savn} the authors propose to use meta-learning for test time adaptation to the environment for efficient handling of the train-test distribution shift. 
\textbf{A3C+MAML} shows a considerable performance boost over the simple \textbf{A3C} baseline as shown in Table \ref{tab:results_compare}. Despite that, when combined with our method in \textbf{A3C+MAML+\sexyname} we observe an additional absolute improvement of around 5\% in success rate on both the short and long trajectories. This further demonstrates the modularity of our approach by showing that it improves both adaptive, and  non-adaptive,  baseline methods. Note that while the absolute improvement in both the short and long trajectories present similar figures, the relative improvement is much larger for long trajectories. This, again, supports the previous hypothesis that our imagination can effectively help address both the LSTM forgetting problem and finding the correct path to the successful target. Furthermore, nearly 2\% improvement compared to \textbf{A3C+MAML+\sexyname} on the SPL over the long trajectories implies that \sexyname helps avoid futile wandering around.

The third baseline method that we compare against in the second setting is \textbf{A3C+ORG}~\cite{tpn_anu}. The main contribution of the authors of this method is to use an off-the-shelf object detector, \eg FasterRCNN \cite{fasterrcnn}, and incorporate the detected object bounding boxes along with confidence scores by building a neural graph, inspired by~\cite{scene_priors}. Their method builds a better state representation that renders finding target objects much easier for the agent. As long as the objects are detected by the object detector the agent can learn to directly navigate towards them taking a relatively short trajectory. 

However, in order to show that our \sexyname module is general enough to work even in conjunction with an object detector we provide the results for \textbf{A3C+ORG+\sexyname}. We observe that our method improves the success rate by over \%2 compared to \cite{tpn_anu}. This shows that, irrespective of the quality of the state representations empowering the agent with imagination, using our \sexyname module can help improve navigation performance. It also confirms the value of imagination for navigation,  and the modularity and generalisability of our simple approach.
Note that we do not further compare our results with the addition of imitation learning into this problem, as the authors did in \cite{tpn_anu}. This is because we are only concerned with adding imagination into RL in this paper. %\todo{not sure if you really need to add this}

\subsection{Ablation Studies}
\vspace{-2mm}
In this section we present experimental results that provide further insight into the contributions of the different components of \sexyname.
% In this section we run our ablation studies on the second environment setup since we achieve the state-of-the-art results.
\begin{figure*}
    \centering
    \includegraphics[width=0.9\textwidth]{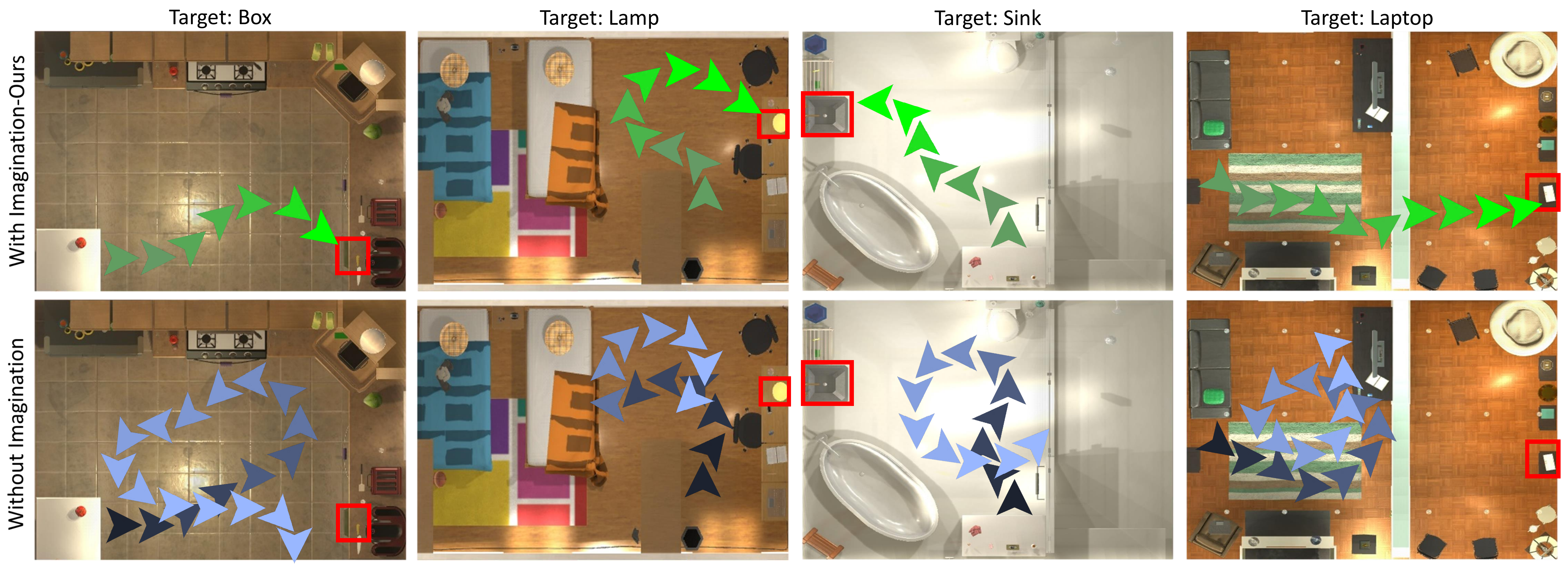}
    \caption{Qualitative trajectory comparison against the baseline method for varying room types and target objects. Our agent can efficiently navigate towards the target by imagining the sub-goal state.}
    \label{fig:trajectory}
\end{figure*}

\begin{figure*}
    \centering
    \includegraphics[width=0.87\textwidth]{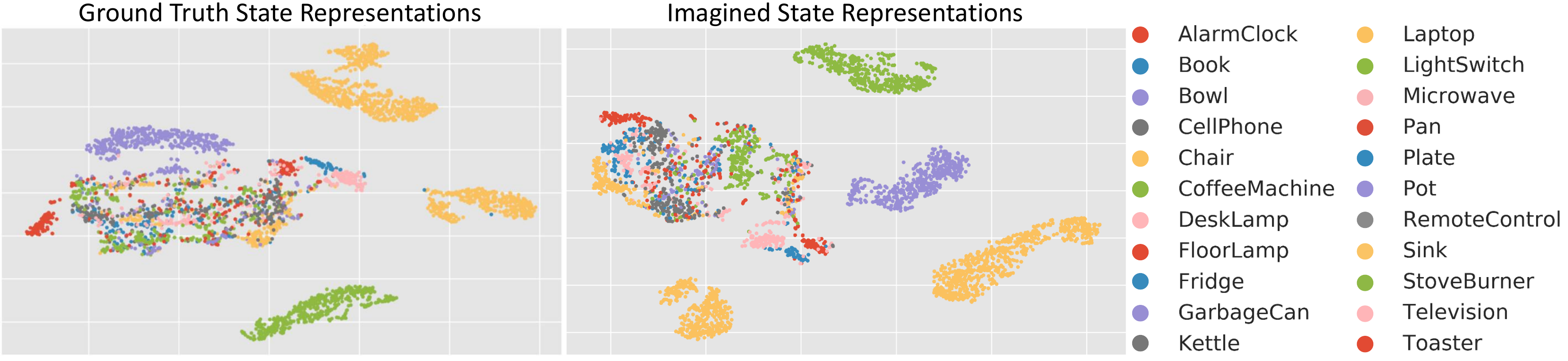}
    \caption{Comparison of the state representations generated using \sexyname  with the ground truth state representations.  \sexyname is conditioned on the target objects so it can learn the structure of the states efficiently.}
    \label{fig:tsne}
    \vspace{-12pt}
\end{figure*}

\begin{table}[b]
%\begin{wraptable}{r}{5cm}
\vspace{-5mm}
\centering
{\small
\begin{tabular}{l|c|c}
\toprule
\textbf{Method}      & \textbf{SPL} & \textbf{SR} \\ \hline \hline
\textbf{A3C+ORG}\cite{tpn_anu}  & 37.5  & 65.3\\
\textbf{Ours-RND} & 37.57       & 64.8 \\
\textbf{Ours-INT} & 37.78       & 63.8 \\
\textbf{Ours-ATT} & 37.76      & 65.4  \\
\textbf{Ours-\sexyname} & \textbf{38.66}       & \textbf{67.6}  \\ \hline
\end{tabular}}
%\caption{Here we respond to the question of "what should the agent imagine?". We can see that using our attention mechanism we can identify the optimal state along the trajectory that its imagination helps with more successful (higher SR) while more efficient navigation (higher SPL).}
\caption{\small{Various ablation studies for our \sexyname.}}\label{tab:ablation}
%\end{wraptable}
% \vspace{-15pt}
\end{table}

\vspace{-15pt}
\paragraph{What to Imagine:} In our approach, we learn to imagine the sub-goal state from the agent's own successful navigation trajectory; we could, however, consider alternative approaches that we compare here. Firstly, we seek to answer whether our \sexyname provides the agent with valuable information about its future sub-goal states. 

%\end{table}
In Table \ref{tab:ablation}, we consider a random imagination by replacing the output of our \sexyname with $\tanh(\bs)$ where $\bs$ is sampled from a Gaussian noise distribution, \ie $\bs\sim\mathcal{N}(\mathbf{0}, 0.5\mathbf{I})$, shown in \textbf{Ours-RND}.  The $\tanh$ non-linearity assures that the noise is similar in value  to the imagined states generated by  \sexyname, but are nonetheless rather random and meaningless. As observed, this leads to a deterioration in the success rate compared to the original baseline, \textbf{A3C+ORG}. We hypothesise that, because it is uninformative, the policy learns to dismiss the randomly imagined input to some extent, but not completely, hence the slight performance degradation.

We further compare our approach with the case of predicting the weighted average of the states in Equation~\ref{eq:att_val}, rather than a single sub-goal. 
The intuition for this experiment is that knowing the attended states gives the agent valuable information about what important future states are expected to be traversed. As shown in \textbf{Ours-ATT} of Table~\ref{tab:ablation} this leads to improved performance which indicates it is helpful to know about the sub-goal states ahead of the agent; however, since the sub-goal is not always identifiable (multiple sub-goals and trajectories could have the same attended states) the performance gain is rather insignificant.

%\todo{I think it is better to change AVG to attention}

% In a third comparison, we compare our method with \textbf{Ours-AVG}. In that method, our \sexyname module is trained on a weighted average of the states along a successful trajectory based on the attention weights in Equation~\ref{eq:att_val}. As can be seen the performance is slightly better than the original baseline. Our explanation is that the attended sum of states still contains information about the important sub-goal state ahead of the agent. However, this information is occluded by other states, as a result of summation, hence the observed performance boost is limited.
\vspace{-13pt}
\paragraph{Imagination Intervals:} Rather than imagining the sub-goal from the initial state, we can consider imagining in multiple intervals, \ie in a fixed interval and number of steps. 
%/A*** That's not at all clear
In that case, in each interval a different sub-goal is predicted. As shown in \textbf{Ours-INT} in Table \ref{tab:ablation}, this leads to a performance degradation. We believe this is because the policy collapses to a deterministic one due to the constraints on predicting multiple sub-goals. Consequently, the agent stops exploring and converges to a sub-optimal solution.

% Our \sexyname can also be trained to re-imagine in given fixed intervals during the trajectory execution, named as \textbf{Ours-INT} in Table \ref{tab:ablation}. In that method, the idea is that the imagination will guide the agent towards a near future state that leads to success. We can see that the performance degrades compared to the baseline method. We hypothesise this method limits the agent's exploration and hence generalisation due to learning to strictly follow previously taken trajectories.

\vspace{-13pt}
\paragraph{Explicitly Structured Imagination}
%\todo{what motivates this? we did not discuss using any sort of autoencoder.}
It can be seen in Figure~\ref{fig:tsne} that \sexyname can accurately learn the structure in the state representations used by the agent's state encoder. To further investigate the quality of the imagined states we train a Conditional Variational Auto-Encoder (C-VAE) \cite{cvae} on the state representations. Then we use  \sexyname  to generate the latent state of the C-VAE rather than working directly on the agent's state representations. We compare the average loss achieved by our module trained directly on the agent's state representations with that of trained on the latent representations of the C-VAE. We observe that direct imagination achieves a lower average loss, $0.012$ vs $0.018$, while being significantly less complex.
%in figure \ref{vae_loss}.

%\begin{wrapfigure}{l}{0.2\textwidth}
%    \includegraphics[width=0.22\textwidth]{latex/vae_loss.png}
%  \caption{Comparing average imagination loss over 4000 episodes using our \sexyname module and C-VAE.}
%  \label{vae_loss}
%\end{wrapfigure}
%\vspace{-6mm}
\subsection{Qualitative Comparison}
In Figure \ref{fig:trajectory} we compare four sample navigation episodes between our method and the baseline method \cite{savn}. It can be commonly seen in all of these trajectories that  \sexyname  enables the agent to navigate towards the target object more intelligently. The baseline method, however, has difficulty finding the correct path and stops after some wandering around.

%\subsection{Quality of Imagined Latent Representations}
%\todo{more explanation}
%Here we use t-SNE \cite{tsne} dimensionality reduction to visualise the states learned by our imagination module and compare it with the ground truth target states.

%\todo{move to experiment:First we need to identify what the \sexyname module should generate, \ie, imagine. A trivial solution is to given a successful trajectory $\tau$ sampled from the agent's policy $\pi$ select the final state as the target state that the agent needs to learn to imagine. The major drawback of this method, however, is that it may bias the policy towards a sub-optimal convergence. The reason is that the agent might have observed a more optimal sub-goal state during the trajectory before stopping at a sub-optimal goal state. We propose a novel way to identify the optimal state during a trajectory as a sub-goal state to train the imagination module using multi-head attention mechanism and thus avoid learning a sub-optimal imagination.}

\section{Conclusion and Future Work}
We have shown that it is possible to empower visual navigation agents with foresight by the addition of \sexyname. This is achieved by imagining a future sub-goal in the latent space that leads to a successful navigation episode. Extensive experiments %and ablation studies
show how that \sexyname
 can be integrated into a wide variety of methods to improve their navigation performance.
While it has proven effective, we consider this work as an initial step towards enabling imagination for visual navigation. As the next steps we plan to investigate the role of imagination enabling better exploration. We also intend to consider learning from failures by updating the imagined sub-goals that did not lead to success. 
{\small
\bibliographystyle{ieee_fullname}
\bibliography{main_paper}
}

\end{document}